\documentclass[letterpaper]{article}
\usepackage{preprint_nac}

\usepackage[utf8]{inputenc} 
\usepackage[T1]{fontenc}    
\usepackage{hyperref}       
\usepackage{url}            
\usepackage{booktabs}       
\usepackage{amsfonts}       
\usepackage{nicefrac}       
\usepackage{microtype}      

\usepackage{booktabs} 
\usepackage{subcaption}
\usepackage{amsmath}
\usepackage{mathtools}
\usepackage[toc,page]{appendix}
\usepackage{enumitem}
\usepackage{graphics}
\usepackage{graphicx}
\usepackage[export]{adjustbox}
\usepackage{algorithm}
\usepackage{algorithmicx}
\usepackage[noend]{algpseudocode}
\usepackage{multirow}

\title{Recognizing and Verifying Mathematical Equations using Multiplicative Differential Neural Units}

\author{%
  Ankur Mali$^\dagger$ \\
  \texttt{aam35@psu.edu} \\
  \And
  Alexander Ororbia$^\ddagger$ \\
  \texttt{ago@cs.rit.edu} \\
  \And
  Daniel Kifer$^\dagger$ \\
  \texttt{duk17@psu.edu} \\
  \And
  C. Lee Giles$^\dagger$ \\
  \texttt{clg20@psu.edu} \\
  \and
  $^\dagger$ The Pennsylvania State University, University Park, PA 16802 \\
  $^\ddagger$ Rochester Institute of Technology, Rochester, NY 14623 \\
  
}

\begin{document}

\maketitle

\begin{abstract}

Automated mathematical reasoning is a challenging problem that requires an agent to learn algebraic patterns that contain long-range dependencies. Two particular tasks that test this type of reasoning are (1) \emph{mathematical equation verification}, which requires determining whether trigonometric and linear algebraic statements are valid identities or not, and (2) \emph{equation completion}, which entails filling in a blank within an expression to make it true. Solving these tasks with deep learning requires that the neural model learn how to manipulate and compose various algebraic symbols, carrying this ability over to previously unseen expressions. Artificial neural networks, including recurrent networks and transformers, struggle to generalize on these kinds of difficult compositional problems, often exhibiting poor extrapolation performance. In contrast, recursive neural networks (recursive-NNs) are, theoretically, capable of achieving better extrapolation due to their tree-like design but are difficult to optimize as the depth of their underlying tree structure increases.
To overcome this issue, we extend recursive-NNs to utilize multiplicative, higher-order synaptic connections and, furthermore, to learn to dynamically control and manipulate an external memory. We argue that this key modification gives the neural system the ability to capture powerful transition functions for each possible input. 
We demonstrate the effectiveness of our proposed higher-order, memory-augmented recursive-NN models on two challenging mathematical equation tasks, showing improved extrapolation, stable performance, and faster convergence. Our models achieve a $1.53$\% average improvement over current state-of-the-art methods in equation verification and achieve a $2.22$\% Top-1 average accuracy and $2.96$\% Top-5 average accuracy for equation completion. 
\end{abstract}

\section{Introduction}
\label{sec:intro}
Mathematical reasoning is one problem domain that crucially requires understanding the composition between symbols and arithmetic operators. In demonstrating that it has an ``understanding'' of basic mathematical and logical concepts, an agent must solve new equations or resolve expressions that might become increasingly more complex with time. This entails understanding the structure of equations, as well as their underlying grammar, in order to properly and effectively extrapolate to unseen examples.
With respect to this kind of reasoning, artificial neural networks (ANNs) have been shown to experience great difficulty achieving the robustness, adaptability, and flexibility exhibited by human agents \cite{fodor1988connectionism}. Specifically, for tasks requiring the ability to compose knowledge, where complex expressions or structures are created by learning and manipulating rules that permit combination and usage of atomic elements of knowledge (such as the operations of addition or multiplication), ANNs struggle to work correctly and reliably. Indeed, it is often argued that ANNs are incapable of learning to perform the compositional actions needed to mathematically reason \cite{fodor1988connectionism}.

Nonetheless, in this paper, we argue that the limitations in an ANN's ability to learn compositionality is due (at least in part) to several limitations in their current structural design. First, ANNs do not possess the right inductive bias (or prior, in the form of structural constraints) that would allow them to more readily and naturally extract the compostionality in various symbolic languages. Second, standard ANNs, even those that are stateful, e.g., recurrent neural networks (RNNs), lack a proper (interpretable) memory structure that would be allow them to properly handle the arrangements of symbols that compose mathematical expressions of increasing depth (complexity) and length. Since mathematical equations are derived from context-free languages related to mathematical identities \cite{arabshahi2018combining}, it would make sense to manipulate an external memory when processing equations. For example, a model trained on $(\sqrt{1} \times 1 \times y) + x = (1 \times y) + x$ should be capable of generally understanding structure that includes equality and inequality. Furthermore, a memory structure would allow a network to better generalize to unseen equations of different depths, since memory can offload some of the memorization that a network typically does using its short-term synapses for \cite{arabshahi2019memory}. For instance, a model augmented with external memory should be capable of understanding the following equation (without any need to train it directly on it): $y \times \Big(1^1 \times (3 + (-1 \times 4^{0 \times 1})) + x^1\Big) = y \times 2^0 \times (2 + x)$. 

Ultimately, we aim to better understand the components required to enhance an ANN's ability to mathematically reason and, therefore, we explore two important reasoning tasks: mathematical equation verification (is a stated identity true?) and mathematical equation completion (fill in the blank to make the equation true). Both of these tasks are designed such that an agent must learn how to combine and arrange symbols and operators in order to properly parse complex mathematical equations like the two provided in the prior paragraph. 
However, most, if not all, modern-day neural architectures are ill-suited to properly tackle these problems, as evidenced by several recent studies in related tasks that require memory acquisition and knowledge of compositionality.
For example, in the domain of semantic parsing, research indicates that as task complexity increases, i.e., sequence length increases and relationships between tokens/items becomes complicated, RNNs fail to generalize to unseen examples \cite{lake2018generalization}.
Similarly, poor generalization was observed when RNNs were trained to grammatically infer complex Dyck languages \cite{mali2019neural, mali2020recognizing,Lstmdynamiccounting} and when neural transformers \cite{vaswani2017attention} were trained to do symbolic integration and solve differential equations \cite{lample2019deep,saxton2019analysing}.

This failure to generalize stems from the fact that models like the transformer are theoretically not capable of recognizing complex grammars \cite{hahn2020theoretical} and, empirically, only perform well when the test set largely comes from the same distribution as the training set \cite{lample2019deep}. In essence, once a distributional shift occurs (test samples begin to vary significantly from training samples), models like the neural transformer break down and fail to extract the functionality of symbols presented to it (let alone learn how to actually compose them). In the realm of mathematical reasoning, such a shift could occur by simply changing the location of an operator (changing its role), or connecting it to different symbols/arguments, or even switching which side of equality it belongs. Modern-day ANNs fail to operate in these kinds of out-of-distribution cases.

Therefore, having the ability to extract and learn relationships between various operators and complex structures/entities/objects (that make up symbolic expressions) would facilitate such out-of-sample generalization when processing mathematical equations. One type of (data) structure that could serve as a useful inductive bias for the ANN is the binary tree, which is a useful way of representing equations. Such trees, upon construction, are interpretable and often inspected when attempting to understand relationships between symbols and operators. When data is available in tree form, recursive neural networks, e.g., tree RNNs, have been shown to outperform standard RNNs \cite{tai2015improved,socher2011parsing,allamanis2017learning,evans2018can,arabshahi2018combining}, generalizing well by exploiting their own natural tree-like design (making them suitable for processing equations).
However, as the depth of their underlying tree increases, recursive networks struggle to generalize to unseen, longer strings, despite their suitable design. One reason for this is that, as tree depth increases, the credit assignment problem becomes more challenging, hampering the system's learning ability. On the other hand, the lack of a proper error-correction mechanism in the model itself (in the event there is no learning) limits its ability to adapt to novel pattern sequences. In the hopes of endowing recursive networks with some form of error-correction, recent work has attempted to combine differentiable memory with tree-based Long Short Term Memory (LSTM) models \cite{arabshahi2019memory}. However, this work is limited since it focuses on adding external memory to a standard neural controller (hoping this alone will improve extrapolation ability), assuming that the original network is already sufficiently good at modeling compositionality instead of addressing the network's potential weaknesses.
Our work will challenge this assumption by making the controller more powerful.  

To fundamentally resolve the limitations described above, in this work we look to creating more powerful recursive networks by generalizing them to utilize higher-order synaptic weight parameters. Higher-order (or tensor) synaptic connections have been shown, both classically and in recent efforts, to be provably capable of encoding complex grammars, especially in the context of temporal neural models that tackle the challenging task of processing and inferring complex context-free grammars \cite{omlin1996constructing, mali2019neural,stogin2020provably}. With tensor parameters, there is a straight forward mapping of a state machine into a set of transition rules that can be programmed into an ANN's representations, something which is currently not possible for first-order models (which is what most modern networks actually are). This, we argue, makes higher-order synapses a potentially useful inductive bias for recursive networks, especially since mathematical equations are a kind of context-free grammar. Furthermore, we account for the higher computational cost imposed by using higher-order parameters in an ANN (which has prevented their widespread use in modern-day deep learning) by developing several approximations based on multiplicative interactions. These approximations are used in tandem with a differentiable stack memory, which prior work has shown is important in allowing RNNs to recognize certain classes of grammars \cite{suzgun2019memoryaugmented, Lstmdynamiccounting, mali2020recognizing}.

The primary contributions of this paper are as follows:
\begin{itemize}
    \item We introduce the first higher-order recursive recurrent neural network, known as the second-order Tree-RNN.
    \item We introduce new variant models that approximate second-order Tree-RNNs.
    \item We introduce new stack-augmented variants of these higher-order recursive recurrent models, beating out state-of-the-art results for two mathematical reasoning tasks.
\end{itemize}

\begin{figure}[t]
    \centering
    \includegraphics[width=0.55\textwidth]{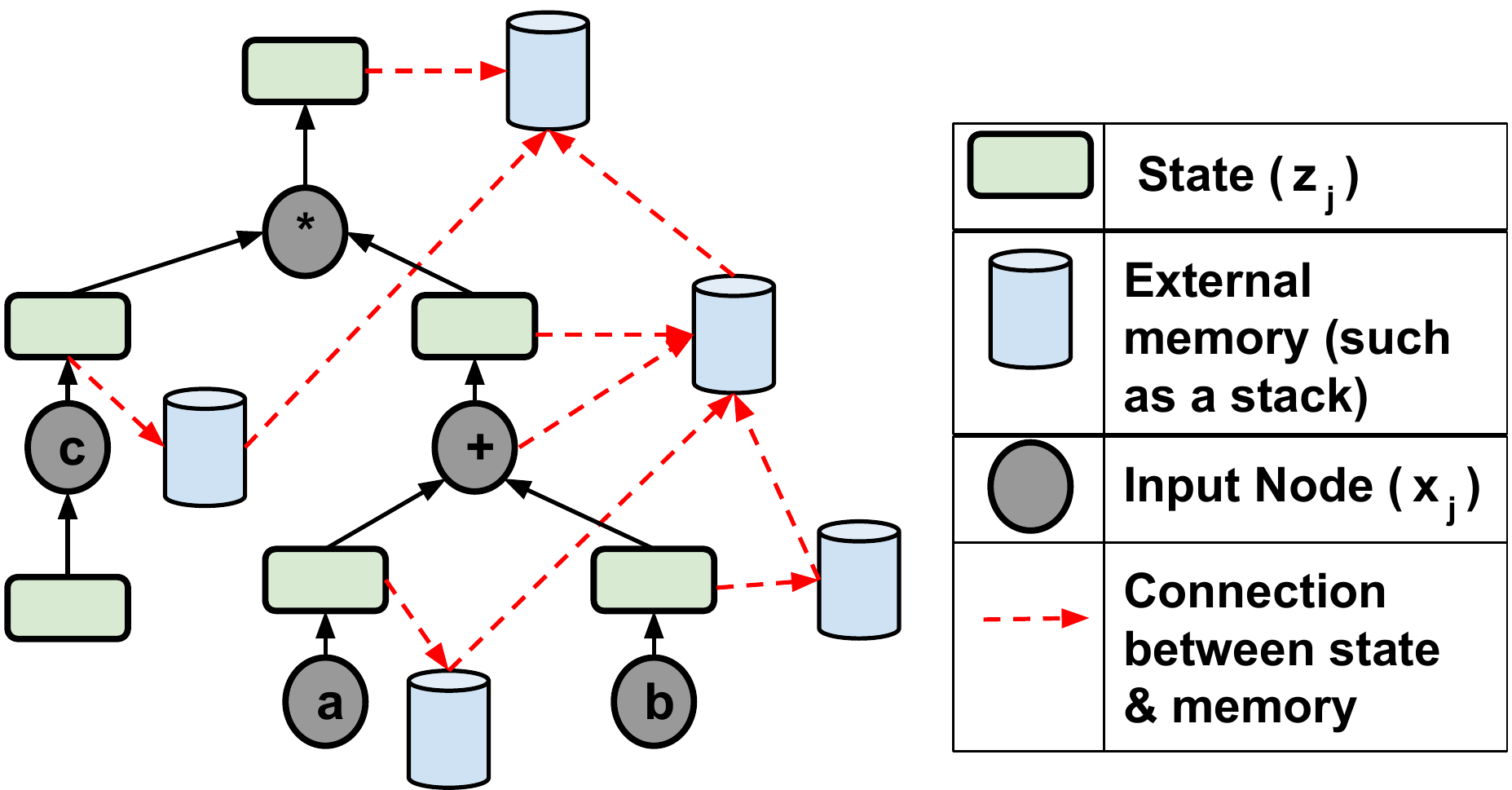}
    \caption{Architecture of the recursive neural network with nodes augmented with external memory. The network shown is operating on the equation $c * (a + b)$ and has arranged its nodes to process the underlying symbol sequence.}
    \label{fig:recursivenn}
    \vspace{-0.5cm}
\end{figure}

\section{Background and Notation}
\label{sec:background}

In this section, we define the notation to be used in this paper.
We denote scalars with non-bold letters (e.g., $c$), vectors with bold lowercase letters (e.g., $\mathbf{x}$), and matrices with bold uppercase letters (e.g., $\mathbf{W}$).
A recursive neural network (recursive-NN) is a kind of tree-structured neural architecture in which each node is represented by an ANN (see Figure \ref{fig:recursivenn}). Such a design has been empirically shown to capture semantic relationships in symbolic data, allowing the model to better generalize to harder problems in natural language processing \cite{tai2015improved,arabshahi2018combining, socher2013recursive,bowman2015recursive}. 
All the nodes of a rec-RNN have a state denoted by $\mathbf{z}_j \in \mathbb{R}^n$ and an input denoted by $\mathbf{x}_j \in \mathbb{R}^{kn}$ where $n$ is the hidden dimension, $k$ is number of children nodes, $N$ is the number of nodes in the tree, and $j \in [0,N-1]$. The children of node $j$ are denoted as $c_{jk}$, where k is number of children for node$j$. The hidden representation $\mathbf{z}_j$ depends on the network architecture, acting as memory for the model. For instance, in a simple Tree RNN, state $\mathbf{z}_j$ is computed by passing an input $\mathbf{x}_j$ through a feedforward network.
However, since recurrence is absent in this memory node, we believe that the model is limited in its ability to transfer knowledge across time steps. Furthermore, other factors we argue hinder model performance are: 1) the absence of a better input encoding, and 2) the fact that neurons in the Tree RNN are trying to both model compositionality and extrapolate across large sequences, creating a harder learning problem than needed. 
To address these issues, we will develop several novel variations of the Tree RNN that integrate recurrent second order synapses (or approximations thereof) and external (stack) memory.

\section{Second-Order Tree Recurrent Networks} %
\label{sec:2nd_order_treenns}
In this section, we describe our second ($2^{\text{nd}}$)-order Tree RNN architecture. This is a tree structure where every node $j$ contains a $2^{\text{nd}}$-order RNN. In a $1^{\text{st}}$-order RNN, including popular variants such as the Long Short-Term Memory LSTM \cite{hochreiter1997long}, the hidden state $\mathbf{z}^{t-1}_j$ of node $j$ at time $t-1$ is combined with its input $\vec{x}_j$ to produce the hidden state at time $t$: $\mathbf{z}_j^t = f(W_1\mathbf{z}^{t-1}_j + W_2\mathbf{x}_j)$, where $f$ is an activation function. In contrast, a  $2^{\text{nd}}$-order RNN uses a third-order weight tensor $\mathbf{W}$ to update the state \cite{giles1990higher,omlin1996constructing} as follows:
\begin{align}
  \mathbf{z}^t_j[\ell] = f\left(\sum_{i_1,i_2}\mathbf{W}[\ell, i_1, i_2] \mathbf{z}^{t-1}_j[i_1] \mathbf{x}_{j}[i_2] + \mathbf{b}_j[\ell]\right) \mbox{.}
\end{align}
This tensor operation can also be written as $\mathbf{z}^t_j = f\left(\mathbf{W}( \mathbf{z}^{t-1}_j, \mathbf{x}_{j}) + \mathbf{b}_j\right)$. The benefit of this update style is that it concisely captures the state transitions of a deterministic finite automata (DFA). For example, if $\mathbf{x}_j$ is a one-hot encoding of an input symbol (say $i_1$, with a vocabulary size $V$) and $\mathbf{z}_j$ is a one-hot encoding of an automaton state (say $i_2$), then $\mathbf{W}[:, i_1, i_2]$ can be set to be the one-hot encoding of the state that the  DFA should transition to. Thus, the $2^{\text{nd}}$-order RNN architecture represents a differentiable transition function over a distributed state and input representation \cite{giles1990higher,mali2019neural}. Tensor operations have also been experimented with ANNs containing feedforward tensor weights \cite{socher2013reasoning}, yielding promising results on logical reasoning data, outperforming first-order networks \cite{bowman2013can, socher2013reasoning}. In contrast to these related efforts, our work explores tensor recurrent weights combined with recursive connections.

To get a $2^{\text{nd}}$-order Tree RNN, we set each node $j$ to be a $2^{\text{nd}}$-order RNN which takes in as input $\mathbf{x}_j$ a concatenation of its $k$ children. Hence, the tensor weight matrix $\mathbf{W}$ has dimensions $n\times n \times kn$. Note that we also set activation $f$ to be the hyperbolic tangent.
To the best of our knowledge, this work is the first to propose a recursive-NN (tree) model built with higher order weights, which, as we will show, is more powerful than a simple Tree-RNN (a Tree-RNN uses feedforward ANNs to parameterize its nodes and hence does not exploit recurrence well).

To effectively handle the processing of equations, we further incorporate hidden-to-output weight transition matrices. 
One can either share weights or have node-specific weight parameters in the architectures. In the case of equations, we assign a separate set of parameters to each unique operator that could be used to define an equation, allowing the rec-RNN to learn relationships across these operators. When an operator or variable is duplicated, i.e., it occurs more than once in an equation symbol stream, we share the parameters across the occurrences. 

\subsection{Multiplicative Tree LSTM} 
\label{sec:mtree_lstm}
Despite having theoretical justification and better expressivity, $2^{nd}$ order networks are computationally expensive, scaling poorly to large datasets due to their higher-order synapses, which create a challenging optimization problem over a large computation graph.
To overcome this limitation , we take inspiration from recent success in approximating higher order synaptic connections. Specifically, we look to RNNs that use multiplicative weights, which shown promise in tasks such as grammatical inference \cite{mali2020recognizing} and sequence modeling \cite{wu2016multiplicative,krause2016multiplicative}.
Multiplicative synapses are inspired by $2^{nd}$ order connections \cite{giles1990higher,giles1993extraction,nndpa1998sun,rabusseau2019connecting} and aim to approximate their computation, having also been shown to provably encode complex grammatical structure \cite{omlin1996constructing, mali2019neural} (yielding more interpretable RNNs that generalize well). A multiplicative network introduces an intermediate state ($\mathbf{m}$) that creates a flexible input-output transition that also endows the RNN with implicit error handling and improved generative ability \cite{wu2016multiplicative}.
Based on this, we design a novel recursive-NN that employs multiplicative synapses instead of second-order tensor weights as a form of memory. Specifically, we propose the multiplicative Tree Long Short-Term Memory (MTree-LSTM) (generalizing the tree-LSTM \cite{tai2015improved}), a type of recursive-NN with nodes made up of multiplicative weights that allow for different transition matrices for each possible input (much like second-order synapses do). 

The above multiplicative memory structure is coupled with LSTM gates to actively process sequences containing long-range dependencies. For an $N$-ary MTree-LSTM , the branching factor of the model's underlying structure is at most $N$, assuming that children nodes are ordered, \emph{i.e.}, we can index them from $1$ to $N$. For any node $j$, the hidden state and memory cell of the $n$th child would then be represented as $\mathbf{z}_{jn}$ and $\mathbf{c}_{jn}$ respectively. 

The resulting $N$-ary MTree-LSTM transition equations, following the description above, would then be:
\begin{align}
\mathbf{\hat{z}}_{n} &= W^{(m)} \mathbf{x}_n + R_z \mathbf{z}^{t-1}_n) \\
\mathbf{m}_{n} &= (W^{(m)}\mathbf{x}_{n} ) +  (R_m \mathbf{\hat{z}}_{n})\\
\mathbf{i}_j &=\sigma \left( W^{(i)} \mathbf{x}_j +  \sum_{n=1}^N U^{(i)}_n \mathbf{m}_{n} + b^{(i)} \right), \label{eq:nary-mtreelstm-first}\\
\mathbf{f}_{jn} &= \sigma\left( W^{(f)} \mathbf{x}_j +  \sum_{n=1}^N U^{(f)}_{n} \mathbf{m}_{n} + b^{(f)} \right), \label{eq:nary-mtreelstm-f}\\
\mathbf{o}_j &= \sigma \left( W^{(o)} \mathbf{x}_j +  \sum_{n=1}^N U^{(o)}_n \mathbf{m}_{n}  + b^{(o)} \right), \\
\mathbf{u}_j &= \tanh\left( W^{(u)} \mathbf{x}_j + \sum_{n=1}^N U^{(u)}_n \mathbf{m}_{n}  + b^{(u)} \right), \\
\mathbf{c}_j &= \mathbf{i}_j \odot \mathbf{u}_j + \sum_{n=1}^N \mathbf{f}_{jn} \odot \mathbf{c}_{jn}, \\
\mathbf{z}_j &= \mathbf{o}_j \odot \tanh(\mathbf{c}_j), \label{eq:nary-mtreelstm-last}
\end{align}
where all input-to-hidden synapses $\mathbf{W}$ are matrices of shape $\mathbb{R}^{n\times 2n}$ and hidden-to-hidden synapses $\mathbf{U}$ are matrices of shape $\mathbb{R}^{2n\times 2n}$. In our experiments, note that $m = z$. Intuitively, we can interpret each parameter matrix as higher-order synapses that encode the correlation between component vectors of the multiplicative unit, the input $x_j$, and the hidden state $z_k$ (obtained from children nodes).

\subsection{Multiplicative-Integration Tree LSTMs}
\label{sec:mi_treelstm}
Much in the same spirit as the MTree-LSTM, in the effort to overcome limitations of $2^{nd}$ order RNNs, we design a multiplicative-integration Tree-LSTM (MI Tree-LSTM), extending the Tree-LSTM \cite{tai2015improved} with nodes that are a first-order approximation of $2^{nd}$ order connections (integrated with LSTM cells). This generally amounts to replacing the addition operation ($+$) with a Hadamard product ($\odot$) in the standard Elman-RNN equation. This element-wise multiplication has been argued to perform a rank-1 approximation of the operation carried by a $2^{nd}$ order connection. These have been shown to be useful in several sequence modeling settings \cite{krause2016multiplicative}.

For an $N$-ary MI Tree-LSTM, the branching factor is at most $N$ (again, children nodes are ordered). For any node $j$, the hidden state and memory cell of the $n$th child is represented as $z_{jn}$ and $c_{jn}$ respectively. Formally, the $N$-ary MI Tree-LSTM transition equations are:
\begin{align}
\mathbf{i}_j &=\sigma \left( W^{(i)} \mathbf{x}_j \odot \sum_{n=1}^N U^{(i)}_n \mathbf{z}_{jn} + b^{(i)} \right), \label{eq:nary-mitreelstm-first}\\
\mathbf{f}_{jn} &= \sigma\left( W^{(f)} \mathbf{x}_j \odot \sum_{n=1}^N U^{(f)}_{n} \mathbf{z}_{jn} + b^{(f)} \right), \label{eq:nary-mitreelstm-f}\\
\mathbf{o}_j &= \sigma \left( W^{(o)} \mathbf{x}_j \odot \sum_{n=1}^N U^{(o)}_n \mathbf{z}_{jn}  + b^{(o)} \right), \\
\mathbf{u}_j &= \tanh\left( W^{(u)} \mathbf{x}_j \odot \sum_{n=1}^N U^{(u)}_n \mathbf{z}_{jn}  + b^{(u)} \right), \\
\mathbf{c}_j &= \mathbf{i}_j \odot \mathbf{u}_j + \sum_{n=1}^N \mathbf{f}_{jn} \odot \mathbf{c}_{jn}, \\
\mathbf{z}_j &= \mathbf{o}_j \odot \tanh(\mathbf{c}_j), \label{eq:nary-mitreelstm-last}
\end{align}
where all the input-to-hidden weights $\mathbf{W}$ and hidden-to-hidden weights $\mathbf{U}$ are matrices in $\mathbb{R}^{n\times 2n}$.  

Note that for both the MTree-LSTM and MITree-LSTM models, the memory cell $\mathbf{c}_j$ is a one-dimensional vector. This cell vector could alternatively be enhanced by instead substituting it with an entire differentiable data/memory structure such as a stack. It is this special situation that we will explore in the next section, where we explicate how to augment our recursive networks with external memory to further increase capacity and improve generalization.

\subsection{Stack-Augmented Tree-RNNs}
\label{sec:stack_treernn}
All of the previously proposed rec-RNN models can be extended to make use of an external, differentiable stack as a means to increase memory capacity. This extension follows in the same spirit as prior research in integrating data structures that improve the generalization ability of RNN models for various sequence processing tasks \cite{joulin2015inferring, mali2020recognizing,suzgun2019memoryaugmented} (though this focused on traditional, first-order RNNs). In our rec-RNN models, each node $j$ is augmented with an external stack $\mathbf{S}_j \in \mathbb{R}^{p\times n}$, where $p$ is the stack size/length.

A stack is a last-in-first-out (LIFO) data structure that an ANN can only interact with by manipulating the structure's top data storage slot. Stacks are often claimed to be more interpretable in nature but, more importantly, they crucially align with formal language theory. Specifically, one key result from theory is that models that make use of two stacks are Turing complete \cite{hopcroft2pda}, meaning that a stack is indispensable when learning context-free languages. When working with mathematical equations, we wish to exploit the computational learning capability that comes with allowing an ANN to control a stack memory, especially given the fact that long-range dependencies are created when extracting structure from equations. 

The top of the stack is denoted by $\mathbf{S}_j[0] \in \mathbb{R}^n$. The stack has two primary operations -- pop and push. Integrating a stack means we are integrating a push-down automation into our network. Specifically, the network will use a $2$D action vector $\mathbf{a}_j \in \mathbb{R}^2$ whose elements represent the push and pop operations for interacting with the stack. These two actions are controlled by the network's state at each node:
\begin{equation}
    \mathbf{a}_j = \phi(\mathbf{A}_j \mathbf{z}_j + \mathbf{b}^{(a)}_j) \label{eq:action}
\end{equation}
where $\mathbf{A}_j \in \mathbb{R}^{2 \times 2n}$ and $\phi$ is the softmax function. We denote the probability of the action ``push'' with the label $\mbox{\emph{actionPush}} = \mathbf{a}_j[0] \in [0,1]$ and ``pop'' with $\mbox{\emph{actionPop}} = \mathbf{a}_j[1] \in [0,1]$. Note that since the softmax nonlinearity has been used -- these two probabilities must sum to $1$ to create a valid distribution over stack actions. 

In order to calculate the next hidden state of our model, we employ the following set of equations:
\begin{align}
\mathbf{\hat{z}_j} = \mathbf{z}_j^{t-1} + \sum^N_{n=1} P \mathbf{S}^{0,t-1}_n,\quad \mathbf{z}_j = f_1(\mathbf{W}_j \mathbf{x}_j \odot R\mathbf{\hat{z}}_j^{t-1})
\end{align}
where we see that a state update for node $j$ is a summation of the top slots of each child node's corresponding stack.
We assume that the top of a stack is located at index $0$. with value $\mathbf{S}_{n}[0]$, via the following:
\begin{multline}
    \mathbf{S}_{n}[0] = \mathbf{a}_j[PUSH]\sigma (D\mathbf{z}_j) + \mathbf{a}_j[POP]\mathbf{S}_{n}^{t-1}[1] \\ + \mathbf{a}_j[NoOP]\mathbf{S}_{n}^{t-1}[0]
\end{multline}
where the symbols PUSH, POP, and NoOP correspond to the unique integer indices $0$, $1$, and $2$ that access the specific action value in their respective slot.
$D$ is a $1 \times m$ matrix and $\sigma(v) = 1/(1 + exp(-v)$ is the logistic sigmoid. If $\mathbf{a}_j[PUSH] = 1$, we add the element to the top of the stack and if $\mathbf{a}_j[POP] = 1$, we remove the element at top of the stack (and shift/move the stack upwards).
Similarly, for the elements stored at depth $i>0$ in the stack, the following rule must be followed:
\begin{align}
    \mathbf{S}_{c_j}[i] = \mathbf{a}_t[PUSH]\mathbf{S}_{c_j}^{t-1}[i-1] + \mathbf{a}_t[POP]\mathbf{S}_{c_j}^{t-1}[i+1] \mbox{.}
\end{align}
Note that for more complex hidden state functions, the state calculation and stack integration is identical to the description in this section. When the above stack is integrated into our previous two models, we obtain the novel variants we call the M-Tree-LSTM+stack and the MITree-LSTM+stack.

\section{Mathematical Reasoning over Equations}
\label{sec:tasks}
In this section, we discuss the tasks investigated, presenting the datasets used for evaluation. Model performance was measured on two challenging benchmark tasks, i.e., mathematical equation verification and equation completion, introduced in \cite{arabshahi2018combining}.

For both tasks investigated, we generated $41,894$ equations of various depths. To create the training/validation/testing splits for this problem, we generate new mathematical identities by performing local random changes to known identities, starting with the $140$ axioms provided by \cite{arabshahi2018combining}. These changes resulted in identities of similar or higher complexity (equal or larger depth), which may be correct or incorrect and are valid expressions within the grammar (CFGs). Models were trained on equations of depths $1$ through $7$ and then tested on equations of depths $8$ through $13$. The data creation process was identical to the one proposed in \cite{arabshahi2019memory}. Table~\ref{tab:data} provides the statistics of the generated samples, showing the number of equations available at each parse tree depth.

\begin{table*}
\small
{%
\begin{tabular}{lcccccccccccccc}
\toprule
\cmidrule(lr){2-15}
 & \bf all & \bf 1 & \bf 2 & \bf 3 & \bf 4 & \bf 5 & \bf 6 & \bf 7 & \bf 8 & \bf 9 & \bf 10 & \bf 11 & \bf 12 & \bf 13 \\
\midrule
\# Eqs & 41,894 & 21 & 355 & 2,542 & 7,508 & 9,442 & 7,957 & 6,146 & 3,634 & 1,999 & 1,124 & 677 & 300 & 189  \\
\addlinespace
CC & 0.56 & 0.52 & 0.57 & 0.62 & 0.61 & 0.58 & 0.56 & 0.54 & 0.52 & 0.52 & 0.49 &  0.50 & 0.50 & 0.50 \\
\bottomrule
\end{tabular}}
\caption{Dataset statistics for the equation sequence dataset (sorted by data depth). CC stands for ``Correct classes''.}
\label{tab:data}
\end{table*}

\noindent \textbf{Data Creation:}
For the equation completion task, we evaluate each model's ability to predict the missing pieces of an equation such that the overall mathematical expression condition holds true. For this experiment, we utilize the same model(s) and baselines used for the mathematical completion task. To create evaluation data, we take all of the generated test equations and randomly choose a node at depth $k$ ($k$ is between $1$ to $13$) in each and every equation. Next, we replace this with all possible configurations for (problem) depth $1$ through $13$ generated using context-free grammars (CFGs) or related generative grammars as suggested by \cite{arabshahi2018combining}. 

Once created, we present this new set of equations to each model/baseline and measure its Top$-1$ and Top-$5$ accuracy (these are reported in the plots found in the main paper). Top$-1$ and Top-$5$ rankings serve as a proxy for each model's confidence when predicting the blank in the mathematical expression (helping us to observe further if the model ensures the correctness of a target expression/equation). Sample equations/expressions from the generated datasets are shown in Table \ref{tab:examples} along with the truth label (the gold standard) and accompanying problem (recursion) depth (which, in turn, serves as a proxy measure of problem difficulty).

\noindent \textbf{Mathematical Equation Verification:}  
In this task, the agent is to process symbolic and numerical mathematical equations from trigonometry and linear algebra. The goal of an agent should be to successfully learn the relationships between these equations and, ultimately, verify their correctness. Verification can be likened to a binary classification problem. Note that we take the equation's parse tree depth as a metric to evaluate hardness of the math problem.


\noindent \textbf{Mathematical Equation Completion:} 
The goal of this task is to predict the missing pieces in a given mathematical equation such that the final outcome holds true (and is mathematically valid). Models trained for equation verification are used to conduct equation completion.  

 

\section{Experiments}
\label{sec:exp}

\noindent
{\bf{Baseline Models:}}
Below we list all of the baselines that we compared to our proposed rec-RNN models, for both of our experimental tasks. The baseline models we compared against were:  {\bf{Majority Class}}: this baseline is a classification approach that always predicts the majority class. 
{\bf{LSTM}}: this baseline is a Long Short-Term Memory network \cite{hochreiter1997long}.
{\bf{Tree-LSTM}}: this baseline is the original Tree-LSTM network \cite{tai2015improved}, where each node is a standard LSTM cell. 
{\bf{Tree-SMU}}: this baseline is the standard recursive neural network coupled with a stack \cite{arabshahi2019memory}.
All of the recursive models (both baselines and proposed models) share parameters whenever the functionality of a given node is similar (in the context of an equation's parse tree). For equation verification, all models optimize the Categorical log likelihood at the output (which is the
root node). The root of the model represents equality and performs a dot product of the output embedding of the right and left sub-trees/nodes. For equation completion, all models are optimized to minimize the cross-entropy loss at the output (the root or equality node). 
The input to the recursive networks includes the terminal nodes (leaves) of the equations -- these terminals consist of symbols (representing variables in the equation) and numbers. The leaves of the recursive networks are embedding layers which encode the symbols and numbers of the equation. 

\noindent
{\bf{Evaluation Metrics:}} For the task of equation verification, we report model performance after measuring accuracy, precision, and recall. We report each metric measurement as a percentage in Table~\ref{tab:verification_results} and \ref{tab:verification_results1} and they are abbreviated as ``Acc'', ``Prec'', and ``Rcl'' for accuracy, precision, and recall, respectively. On the other hand, for equation completion, we report Top-$K$ accuracy \cite{arabshahi2018combining}. Measuring Top-$K$ accuracy is a useful performance metric since it accounts for the percentage of samples for which any given model predicts correctly on at least one correct sample for a given blank. 

\subsection{Implementation Details}
\label{sec:implementation}
All of the models experimented with in this paper were implemented using the PyTorch Python framework \cite{paszke2019pytorch}. Models were optimized using back-propagation of errors to calculate parameter gradients and were updated using the Adam \cite{kingma2014adam} adaptive learning rate, with $\beta_1=0.9$, $\beta_2=0.999$, and by starting its global learning rate at $\lambda = 0.1$ and then employing a patience scheduling that divided this rate by half whenever there was no improvement observed on the validation set. We regularized the models with a weight decay of $0.00002$. 

The number of neurons in each model's hidden layer as well as the drop-out rate were tuned using a coarse grid search, i.e., hidden layer size was searched over the array $[8,15,25,30,40,45,50,55,60,80,100]$ and the drop-out rate was searched over the array $[0.1,0.2,0.3]$. Parameter gradients were estimated over mini-batches of size $50$ for all experiments. We ran all models using $10$ different seeds and report the $10$-trial average and standard deviation of the results. Models were trained for a maximum of $500$ epochs or until convergence was reached, i.e., early stopping was used. The best accuracy of the model is reported when it reaches its best measured performance on the validation set.

\begin{figure}[htb!]
    \centering
    \includegraphics[width=0.8\textwidth]{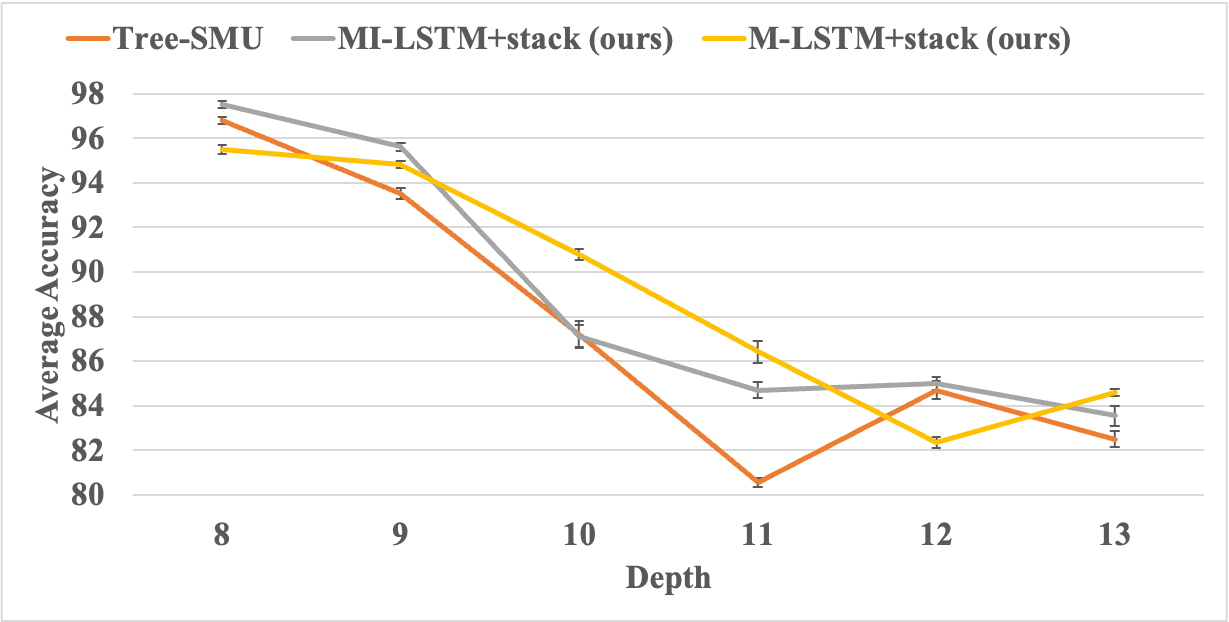}
    \caption{Average accuracy metrics breakdown in terms of performance on the test data depth for equation verification.}
    \label{fig:verify}
\end{figure}
\begin{table*}[htb!]
  
\end{table*}

\begin{table*}[hbt!]

{%
\centering
\begin{tabular}{lcccccc}
\toprule 
\multirow{2}{*}{\bf Approach} & 
	\multicolumn{3}{c}{\bf Train (Depths 1-7)} & 
	\multicolumn{3}{c}{\bf Validation (Depths 1-7)}  \\
\cmidrule(lr){2-4}
\cmidrule(lr){5-7}
 & \bf Acc & \bf Prec & \bf Rcl & \bf Acc & \bf Prec & \bf Rcl \\
\midrule
Majority Class &  58.12 & - & - & 56.67 & - & -
 \\
\addlinespace
RNN \cite{arabshahi2019memory} & 68.50 & 69.61 & 81.27 & 65.77$\pm{0.44}$ & 65.69$\pm{0.72}$ & 83.06$\pm{1.24}$
 \\
LSTM \cite{arabshahi2019memory} & 90.03 & 87.02 & 97.37 & 85.47$\pm{0.27}$ & 81.97$\pm{0.38}$ & 95.32$\pm{0.17}$
 \\

Stack-RNN & 92.03 & 86.42 & 98.43 & 83.47$\pm{0.35}$ & 84.58$\pm{0.32}$ & 95.00$\pm{0.27}$
\\
\addlinespace
Tree-RNN \cite{arabshahi2019memory} & 94.98 & 94.25 & 97.29 & $89.27 \pm{0.04}$  & $87.8\pm{0.39}$ & $94.16\pm{0.61}$
 \\
Tree-LSTM \cite{arabshahi2019memory} & 98.51 & 97.67 & 99.83 & $93.77\pm{0.02}$ & $90.92\pm{0.08}$ & $98.88\pm{0.08}$
\\
\addlinespace

\addlinespace
Tree-SMU & 96.71 & 95.12 & 99.09 & $92.59\pm{0.03}$ & $90.55\pm{0.17}$ &
$98.07\pm{0.18}$ \\
Tree-SMU - no-op & 98.02 & 97.07 & 98.99 & $93.58\pm{0.21}$  & $91.44\pm{0.23}$ & $98.11\pm{0.09}$
\\
Tree-SMU - no-op - normalize & 97.97 &  96.57 & 99.79 & $93.21\pm{0.20}$ &$90.29\pm{0.17}$ & $98.30\pm{0.12}$ 
 \\
\addlinespace
$2^{nd}$ order Tree-RNN (\textbf{ours}) & 95.62 & 94.58 & 98.25 & $90.28 \pm{0.07}$  & $90.25\pm{0.35}$ & $98.08\pm{0.20}$ \\
MI-Tree-LSTM  (\textbf{ours}) & 98.80 & 98.01 & 99.80 & $94.20\pm{0.02}$ & $91.50\pm{0.07}$ & $98.99\pm{0.06}$ \\
MTree-LSTM  (\textbf{ours}) & 98.25 & 97.00 & 98.81 & $94.09\pm{0.02}$ & $91.00\pm{0.05}$ & $99.00\pm{0.06}$
 \\
MI-Tree-LSTM + stack  (\textbf{ours}) & 96.28 & 96.00 & 99.25 & $93.99\pm{0.04}$ & $91.55\pm{0.15}$ & $98.57\pm{0.09}$ \\
M-Tree-LSTM + stack  (\textbf{ours}) & 97.35 & 97.15 & 99.20 & $94.29\pm{0.02}$ & $90.99\pm{0.04}$ & $98.28\pm{0.09}$ \\
\bottomrule
\end{tabular}

\caption{\label{tab:verification_results} Overall performance of the models on train and validation datasets for the equation verification task.}
}

\end{table*}

\begin{table*}[hbt!]

{%
\centering
\begin{tabular}{lcccccc}
\toprule 
\multirow{2}{*}{\bf Approach} & 
	\multicolumn{3}{c}{\bf Train (Depths 1-7)} & 
    \multicolumn{3}{c}{\bf Test (Depths 8-13)} \\
\cmidrule(lr){2-4}
\cmidrule(lr){5-7}
 & \bf Acc & \bf Prec & \bf Rcl  & \bf Acc & \bf Prec & \bf Rcl \\
\midrule
Majority Class &  58.12 & - & - & 51.71 & - & - \\
\addlinespace
RNN \cite{arabshahi2019memory} & 68.50 & 69.61 & 81.27
& 55.5$\pm{0.25}$ & 55.85$\pm{0.61}$ & 67.32$\pm{3.62}$ \\
LSTM \cite{arabshahi2019memory} & 90.03 & 87.02 & 97.37
& 73.09$\pm{0.64}$ & 73.92$\pm{1.48}$ & 74.34$\pm{1.53}$ \\

Stack-RNN & 92.03 & 86.42 & 98.43 
& 73.09$\pm{0.64}$ & 73.92$\pm{1.48}$ & 74.34$\pm{1.53}$ \\
\addlinespace
Tree-RNN \cite{arabshahi2019memory} & 94.98 & 94.25 & 97.29 
& $81.82\pm{0.12}$ & $82.66\pm{0.55}$ & $82.08\pm{0.55}$ \\
Tree-LSTM \cite{arabshahi2019memory} & 98.51 & 97.67 & 99.83
 & $86.8\pm{0.6}$ & $83.68\pm{0.63}$ & $92.54\pm{0.76}$ \\
\addlinespace

\addlinespace
Tree-SMU & 96.71 & 95.12 & 99.09  & ${87.51\pm{0.49}}$ & $84.00\pm{0.31}$ & ${94.21\pm{0.62}}$ \\
Tree-SMU - no-op & 98.02 & 97.07 & 98.99 
 & $87.08\pm{0.15}$ & ${84.32\pm{0.52}}$ & $92.51\pm{0.51}$\\
Tree-SMU - no-op - normalize & 97.97 &  96.57 & 99.79  
  & $87.01\pm{0.50}$ & $83.01\pm{0.62}$ & $93.77\pm{0.49}$ \\
\addlinespace
$2^{nd}$ order Tree-RNN (\textbf{ours}) & 95.62 & 94.58 & 98.25 
& $86.05\pm{0.11}$ & $84.05\pm{0.30}$ & $88.99\pm{0.30}$ \\
MI-Tree-LSTM  (\textbf{ours}) & 98.80 & 98.01 & 99.80 
 & $87.80\pm{0.7}$ & $84.08\pm{0.45}$ & $93.00\pm{0.40}$ \\
MTree-LSTM  (\textbf{ours}) & 98.25 & 97.00 & 98.81 
 & $87.25\pm{0.5}$ & $84.29\pm{0.25}$ & $93.02\pm{0.35}$ \\
MI-Tree-LSTM + stack  (\textbf{ours}) & 96.28 & 96.00 & 99.25 
 & $88.25\pm{0.81}$ & $84.41\pm{0.39}$ & $\mathbf{94.47\pm{0.61}}$ \\
M-Tree-LSTM + stack  (\textbf{ours}) & 97.35 & 97.15 & 99.20
 & $\mathbf{89.04\pm{0.54}}$ & $\mathbf{84.59\pm{0.47}}$ & $94.39\pm{0.59}$ \\
\bottomrule
\end{tabular}
}
\caption{\label{tab:verification_results1} Overall performance of the models on train and test datasets for the equation verification task.}
\end{table*}
\begin{figure}
    \centering
    \begin{subfigure}[b]{0.475\linewidth}
    \includegraphics[width=\textwidth]{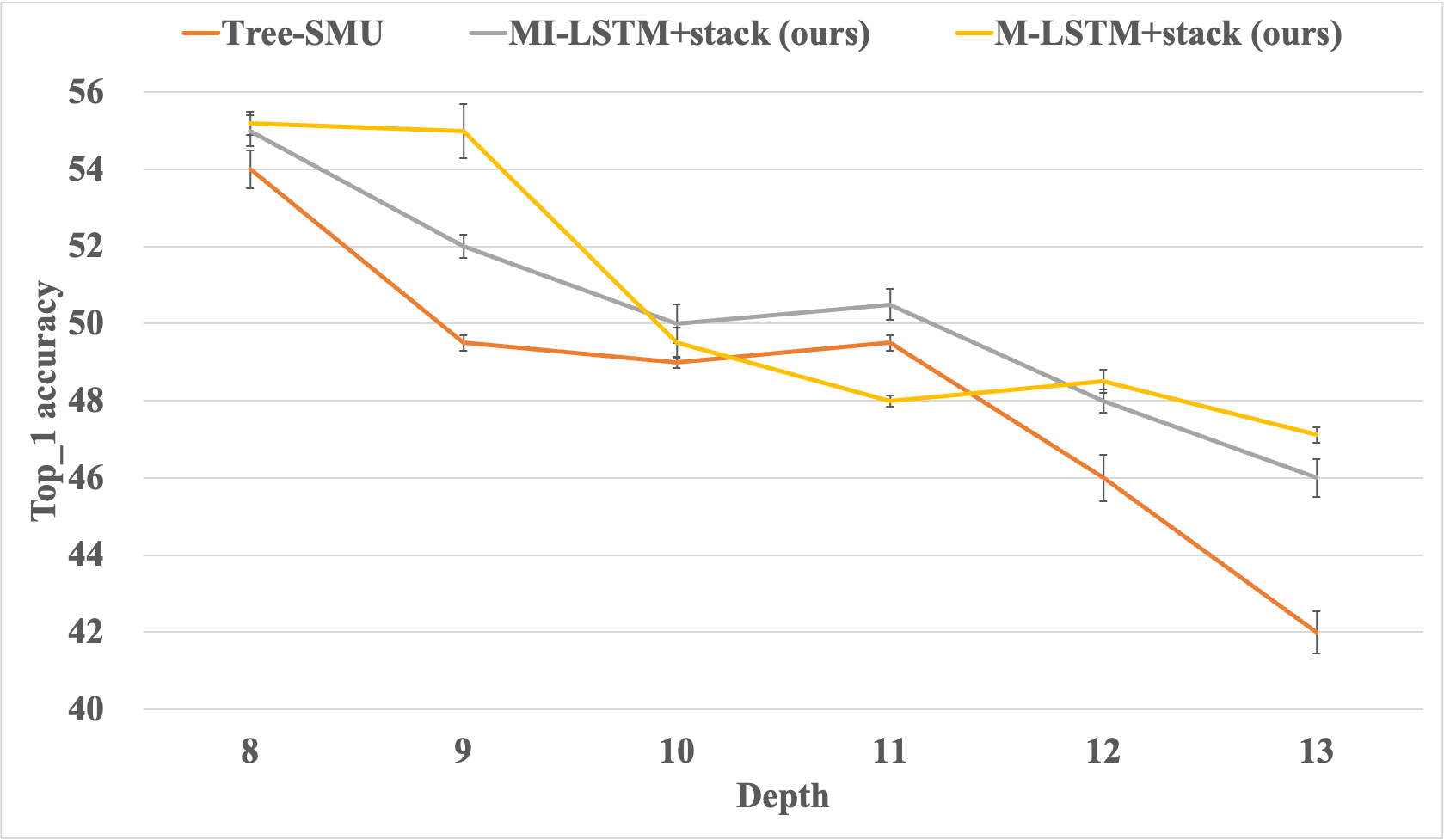}
    \caption{Top 1 Accuracy }
    \label{fig:top1}
    \end{subfigure}
    \begin{subfigure}[b]{0.475\linewidth}
    \includegraphics[width=\textwidth]{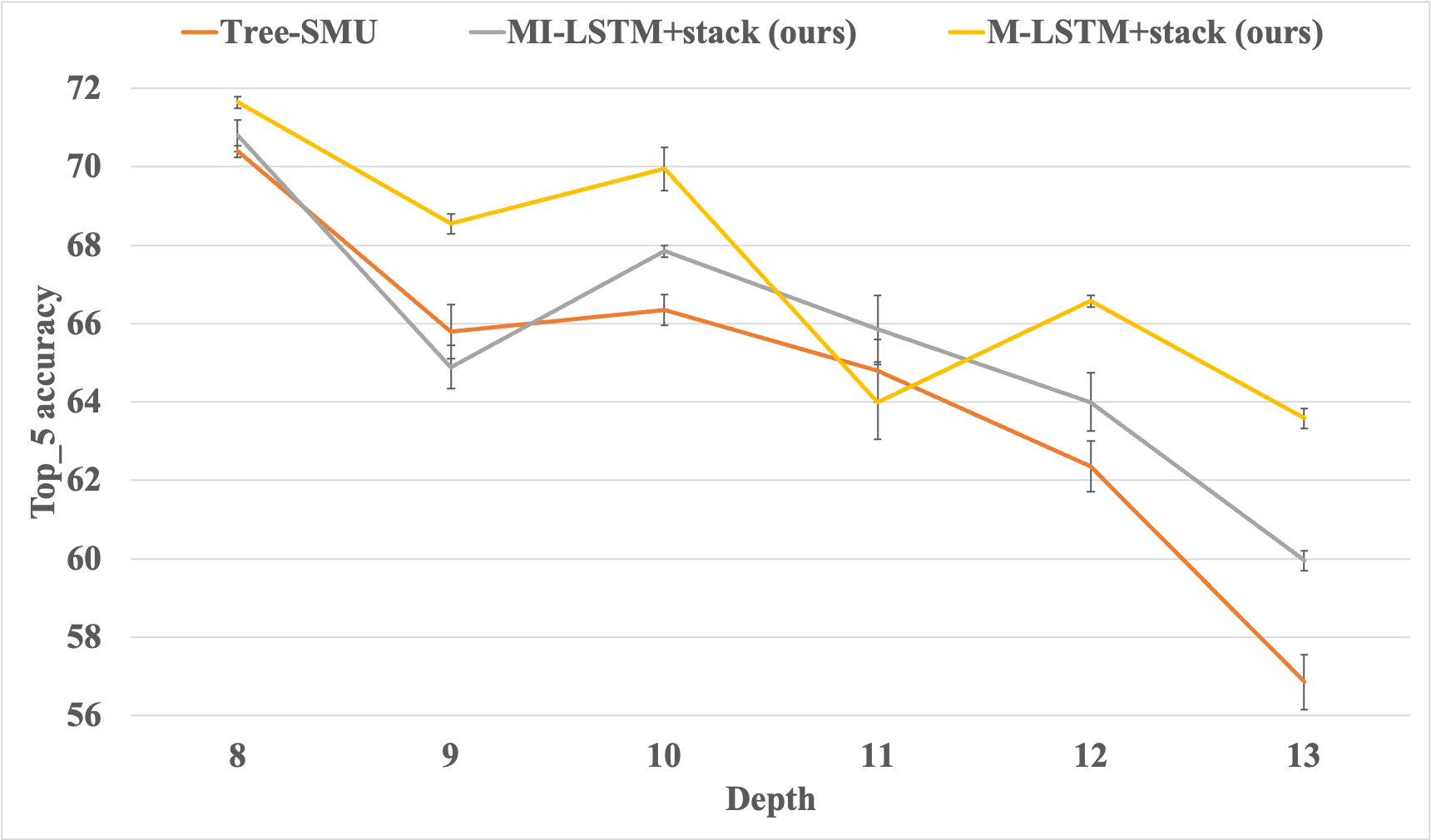}
    \caption{Top 5 Accuracy}
    \label{fig:top5}
    \end{subfigure}
    \caption{Top-$K$ accuracy metrics breakdown in terms of the test data depth for the equation completion task.}
    \label{fig:topk}
\end{figure}

\begin{table}[ht]
\centering
  \begin{tabular}{l | c c c } 
  \hline
  \textbf{Models} & $2$nd Order & $1$st Order  \\
  \hline
  MTree-LSTM  & $87.25$ & $84.59$ \\
  MITree-LSTM  & $87.80$ & $84.59$ \\
  MI-Tree-LSTM+ stack & $88.25$ & $86.51$  \\
  M-Tree-LSTM + stack & $89.04$ & $86.88$ \\
 \hline
 \end{tabular}
  \caption{\label{results:generalization} Average number of epochs (\& validation error reached) required to reach convergence when optimized.}
  \vspace{0.25cm}
  \begin{tabular}{l | c c c } 
  \hline
  \textbf{Models} & Average Epochs & Valid Error \\
  \hline
  TreeLSTM  & $135$ & $93.77\pm{0.02}$ \\ 
  Tree-SMU  & $99$ & $93.58 \pm{0.21}$\\
  MTree-LSTM & $102$ & $94.09 \pm{0.02}$ \\
  MI-Tree-LSTM+ stack & $95$ & $93.99 \pm{0.04}$ \\
  M-Tree-LSTM + stack & $91$ & $94.29 \pm{0.02}$ \\
  $2^{nd}$ order Tree-RNN & $98$ & $90.28 \pm{0.07}$ \\
 \hline
 \end{tabular}
  \caption{\label{results:higher_order} Model test performance (accuracy) with ($2$nd Order) and without higher order ($1$st Order) synapses.}

\end{table}

\subsection{Ablation Study}
\label{sec:ablation}
To complement our results, we provide a small ablation study to demonstrate that the removal of the higher-order synaptic connections leads to a degradation in model performance of the model. This means that the complexity of the synaptic parameters matters in order to achieve consistent performance. As shown in Table \ref{results:higher_order}, we compare recursive neural models with and without higher order synapses. For all the models trained without higher-order weights, the standard LSTM cell was used as memory.  Again, we observe (in tandem with the results presented above) that utilizing second-order weights help yield stable results across problem recursion depths. 

Second, it is important to test whether or not the higher-order parameters offer any computational advantage. To test this, we conducted an experimental analysis of each model's convergence performance (including baselines and proposed architectures). Table \ref{results:generalization} contains the results of this model convergence analysis. All models/baselines were trained $10$ times -- we report the number of epochs that each model required in order to reach best its measured validation accuracy. From the table, it is encouraging to see that the higher-order synapses do indeed result in faster convergence without affecting performance.

Finally, we present extended results in Table \ref{tab:verification_results} and \ref{tab:verification_results1} where the performance of standard RNNs, with and without an external stack memory, are compared to the various recursive networks experimented with earlier. This highlights how difficult the mathematical reasoning task is in general. As discussed before, a recursive structure is quite necessary to solve such complex tasks/problems.
\begin{table*}[htb!]
\centering
\resizebox{1\textwidth}{!}{%
\begin{tabular}{llc}
\toprule
\bf Example & \bf Label & \bf Depth \\ 
\midrule
\addlinespace
$(\sqrt{1} \times 1 \times y) + x = (1 \times y) + x$ & Correct & 4 \\
\addlinespace
$\sec(x + \pi) = (-1 \times \sec(\sec(x)))$ & Incorrect & 4 \\
\addlinespace
$y \times \Big(1^1 \times (3 + (-1 \times 4^{0 \times 1})) + x^1\Big) = y \times 2^0 \times (2 + x)$ & Correct & 8 \\
\addlinespace
$\sqrt{1 + (-1 \times (\cos(y + x))^{\sqrt{\csc(2)}} )} \times (\cos(y + x))^{-1} = \tan(y^1 + x)$ & Incorrect & 8\\
\addlinespace
$2^{-1} + \Big(-\frac{1}{2} \times \big (-1 \times \sqrt{1 + (-1 \times \sin^2(\sqrt{4} \times (\pi + (x \times -1))))} \big) \Big) + \cos^{\sqrt{4}}(x) = 1$ & Correct & 13 \\
\addlinespace
 $\big(\cos(y^1 + x) + z\big)^w = \big(\cos(x) \times \cos(0 + y) + \big(-1 \times \sqrt{1 + -1 \times \cos^{2}(y + 2\pi)}\; \big) \times \sin(x) + z \big)^w$  & Correct & 13  \\
 \addlinespace
 $\sin\Big( \sqrt{4}\;^{-1} \pi + (-1 \times \sec \big( \csc^2(x)^{-1} + \sin^2(1 + (-1 \times 1) + x + 2^{-1} \pi ) \big) \times x) \Big) = \cos(0 + x)$ & Incorrect & 13 \\
\addlinespace
\bottomrule
\end{tabular}}
\caption{\label{tab:examples} Examples of generated equations used in the paper's experiments \cite{arabshahi2019memory}.}
\end{table*}
\section{Results and Discussion}
\label{sec:results}

\paragraph{Equation Verification:} In Figure \ref{fig:verify}, we compare our best model with the recently proposed state-of-the-art model in \cite{arabshahi2019memory}. Of the five proposed models, as shown in Figure \ref{fig:verify}, the M-Tree-LSTM+stack and the MI-Tree-LSTM+stack reach the best performance and are stable across equation examples (as the depth of the test set is increased).
In Table \ref{tab:verification_results} and \ref{tab:verification_results1}, we provide a more extensive performance comparison across all models, baselines, and the proposed rec-RNNs on the equation verification task. 

\paragraph{Equation Completion:} To further test the generalization ability of the various models studied in this paper, we next turned our attention to the equation completion task. 
In Figures~\ref{fig:top1} and \ref{fig:top5}, we report the top-$1$ and top-$5$ accuracy measurements of various recursive models. Notably, observe that the performance of the proposed higher-order rec-RNNs is consistently better than that reported in prior work for both models with and without differentiable memory. The performance of standard RNN models on this task was quite poor as evident in table \ref{tab:verification_results}  and \ref{tab:verification_results1}. This strongly demonstrates that just simple (first-order) recurrent architectures fail to capture much, if any, useful compositional knowledge underlying mathematical equations.


\section{Related Work}
\label{sec:related_work}
The mathematical reasoning problems considered in this work are examples of neural programming, or a task family that requires an ANN to learn (complex) structures such as programs, mathematical equations, and logic from data~\cite{allamanis2017learning,evans2018can,graves2014neural, zaremba2014learning,reed2015neural,cai2017making,saxton2019analysing}. Neural programming tasks are a key application for testing an ANN's ability to extrapolate and compose elements of knowledge. Grammatical inference has also often served as another kind of neural programming problem that challenges an RNN's ability to extract useful hierarchical representations over symbolic sequences, especially as the grammar complexity increases \cite{suzgun2019memoryaugmented,Lstmdynamiccounting,mali2020recognizing}.

Recursive networks, which theoretically could prove invaluable for neural programming, have been used to model a wide of compositional data types across many applications (all of which contain an inherent hierarchy nested in the data) \cite{hupkes2018visualisation, hupkes2020compositionality}, e.g., natural scene classification \cite{socher2011parsing}, sentiment classification, semantic relatedness/syntactic parsing  \cite{tai2015improved,socher2011parsing, socher2013reasoning,socher2013recursive,bowman2013can,bowman2015recursive}, and neural programming and logic \cite{allamanis2017learning,zaremba2014learning, evans2018can}.  
While a great deal of recent research has strived to integrate differentiable memory into standard RNNs \cite{graves2014neural,weston2015, grefenstette2015learning, joulin2015inferring, mali2020recognizing,weston2015, sukhbaatar2015end,graves2014neural,graves2016hybrid,reed2015neural,cai2017making,graves2014neural, das1992learning, kumar2016ask,sun2017neural,mali2019neural,mali2020recognizing}, even based on the theoretical grounding of formal language \cite{hopcroft2pda}, far less work exists on integrating external memory into recursive networks. One notable effort was made in \cite{arabshahi2019memory}, where a recursive network was combined with a stack and LSTM gates. 

Despite the amount of effort that has been spent on augmenting RNNs with memory, to the best of our knowledge, there has been no attempt at designing and generalizing external memory as a means to increase the compositionality abilities of ANNs, especially with the goal of design recursive networks with recurrent weights that better extrapolate to harder problem instances that are often out-of-sample.
Another different, yet related, line of work is that on graph memory networks and tree memory networks \cite{pham2018graph, fernando2018tree} -- however, while powerful models, this differs from our work given that these studies do not investigate the value of higher order connections, approach memory construction differently, and ultimately examine different applications.

\section{Conclusion}
In this paper, we proposed five novel types of recursive recurrent neural networks, which we have shown are useful for modeling compositional data, specifically for processing mathematical equations and expressions. We demonstrated that the performance of most kinds of recurrent networks degrades significantly when the depth or complexity of an input equation increases. 
By generalizing recursive neural networks to use higher-order synaptic connections and to interactively manipulate a stack memory, we designed agents that are capable of acquiring rich, compositional representations of mathematical equations, allowing for out-of-sample generalization. 
More importantly, our work demonstrates that higher-order recursive models consistently achieve stable and overall better performance compared to state-of-the-art baselines for two important mathematical reasoning tasks. 

\bibliographystyle{acm}
\bibliography{ref}

\end{document}


\maketitle

\begin{table}[h]
    \centering
    \caption{Model test performance (accuracy) with ($2$nd Order) and without higher order ($1$st Order) synapses.}
 \label{results:higher_order}
  \begin{tabular}{l | c c c } 
  \hline
  \textbf{Models} & $2$nd Order & $1$st Order  \\
  \hline
  MTree-LSTM  & $87.25$ & $84.59$ \\
  MITree-LSTM  & $87.80$ & $84.59$ \\
  MI-Tree-LSTM+ stack & $88.25$ & $86.88$  \\
  M-Tree-LSTM + stack & $89.04$ & $86.88$ \\
 \hline
 \end{tabular}
 \vspace{0.25cm}
  \caption{Average number of epochs (\& validation error reached) required to reach convergence when optimized.}
  \label{results:generalization}
  \begin{tabular}{l | c c c } 
  \hline
  \textbf{Models} & Average Epochs & Valid Error \\
  \hline
  TreeLSTM  & $135$ & $93.77\pm{0.02}$ \\ 
  Tree-SMU  & $99$ & $93.58 \pm{0.21}$\\
  MTree-LSTM & $102$ & $94.09 \pm{0.02}$ \\
  MI-Tree-LSTM+ stack & $95$ & $93.99 \pm{0.04}$ \\
  M-Tree-LSTM + stack & $91$ & $94.29 \pm{0.02}$ \\
  $2^{nd}$ order Tree-RNN & $98$ & $90.28 \pm{0.07}$ \\
 \hline
 \end{tabular}
 \vspace{-0.7cm}
\end{table}

\subsection{Ablation Study}
\label{sec:ablation}
In this section, we demonstrate that the removal of the higher-order synaptic connections leads to a degradation in model performance of the model, meaning that the complexity of the synaptic parameters matters in order to achieve consistent performance. As shown in Table \ref{results:higher_order}, we compare recursive neural models with and without higher order synapses. For all of the models trained without higher-order weights, the standard LSTM cell was used as memory.  Again, we observe (in tandem with the results presented in the main paper) that utilizing $2$nd-order weights help yield stable results across problem recursive depths. 

Second, it is important to test whether or not the higher-order parameters offer any computational advantage. To test this, we conducted an experimental analysis of each model's convergence performance (including baselines and proposed architectures). Table \ref{results:generalization} contains the results of this model convergence analysis. We trained all of the models/baselines $10$ times and report the number of epochs that each model required in order to reach best measured validation accuracy. From the table, it is encouraging to see that the higher-order synapses do indeed result in faster convergence without affecting performance.

Finally, we present extended results in Table \ref{tab:verification_results} where the performance of standard RNNs, with and without an external stack memory, are compared to the various recursive networks experimented with in the main paper. This highlights how difficult the mathematical reasoning task is in general. As discussed in the main paper, a recursive structure is quite necessary to solve such complex tasks/problems.

\section*{Experimental Details: Data Creation}
\label{sec:details}
For the equation completion task, we evaluate the capability of any model to predict the missing pieces of an equation such that the overall mathematical expression condition holds true. For this experiment, we utilize the same model(s) and baselines used for the mathematical completion task. To evaluate model performance we take all of the generated test equations and randomly choose a node of depth $k$ ($k$ can be anything between $1$ to $13$) in each and every equation. Next, we replace this with all possible configurations for (problem) depth $1$ through $13$ generated using context-free grammars (CFGs) or related generative grammars as suggested by \cite{arabshahi2018combining}. 

Once created, we present this new set of equations to each model/baseline and measure its Top$-1$ and Top-$5$ accuracy (these are reported in the plots found in the main paper). top-1 and Top-5 rankings serve as a proxy for each model's confidence when predicting the blank in the mathematical expression (which helps to dig a bit further into observing if the model understands how to ensure correctness of a target expression/equation). Sample equations/expressions from the generated datasets are shown in Table \ref{tab:examples} along with the truth label (the gold standard) and accompanying problem (recursion) depth (which, as discussed in the main paper, serves as a proxy measure of problem difficulty/complexity).

\begin{table*}[t] 
  \caption{\bf Overall performance of the models on the train and test datasets for the equation verification task.}
\label{tab:verification_results}
\resizebox{1\textwidth}{!}{%
\centering
\begin{tabular}{lccccccccc}
\toprule 
\multirow{2}{*}{\bf Approach} & 
	\multicolumn{3}{c}{\bf Train (Depths 1-7)} & 
	\multicolumn{3}{c}{\bf validation (Depths 1-7)} & 
    \multicolumn{3}{c}{\bf Test (Depths 8-13)} \\
\cmidrule(lr){2-4}
\cmidrule(lr){5-7}
\cmidrule(lr){8-10}
 & \bf Acc & \bf Prec & \bf Rcl & \bf Acc & \bf Prec & \bf Rcl & \bf Acc & \bf Prec & \bf Rcl \\
\midrule
Majority Class &  58.12 & - & - & 56.67 & - & -
& 51.71 & - & - \\
\addlinespace
RNN \cite{arabshahi2019memory} & 68.50 & 69.61 & 81.27 & 65.77$\pm{0.44}$ & 65.69$\pm{0.72}$ & 83.06$\pm{1.24}$
& 55.5$\pm{0.25}$ & 55.85$\pm{0.61}$ & 67.32$\pm{3.62}$ \\
LSTM \cite{arabshahi2019memory} & 90.03 & 87.02 & 97.37 & 85.47$\pm{0.27}$ & 81.97$\pm{0.38}$ & 95.32$\pm{0.17}$
& 73.09$\pm{0.64}$ & 73.92$\pm{1.48}$ & 74.34$\pm{1.53}$ \\

Stack-RNN & 92.03 & 86.42 & 98.43 & 83.47$\pm{0.35}$ & 84.58$\pm{0.32}$ & 95.00$\pm{0.27}$
& 73.09$\pm{0.64}$ & 73.92$\pm{1.48}$ & 74.34$\pm{1.53}$ \\
\addlinespace
Tree-RNN \cite{arabshahi2019memory} & 94.98 & 94.25 & 97.29 & $89.27 \pm{0.04}$  & $87.8\pm{0.39}$ & $94.16\pm{0.61}$
& $81.82\pm{0.12}$ & $82.66\pm{0.55}$ & $82.08\pm{0.55}$ \\
Tree-LSTM \cite{arabshahi2019memory} & 98.51 & 97.67 & 99.83 & $93.77\pm{0.02}$ & $90.92\pm{0.08}$ & $98.88\pm{0.08}$
 & $86.8\pm{0.6}$ & $83.68\pm{0.63}$ & $92.54\pm{0.76}$ \\
\addlinespace
Tree-SMU - no-op - normalize & 97.97 &  96.57 & 99.79 & $93.21\pm{0.20}$ &$90.29\pm{0.17}$ & $98.30\pm{0.12}$ 
  & $87.01\pm{0.50}$ & $83.01\pm{0.62}$ & $93.77\pm{0.49}$ \\
Tree-SMU & 96.71 & 95.12 & 99.09 & $92.59\pm{0.03}$ & $90.55\pm{0.17}$ &
$98.07\pm{0.18}$ & ${87.51\pm{0.49}}$ & $84.00\pm{0.31}$ & ${94.21\pm{0.62}}$ \\
Tree-SMU - no-op & 98.02 & 97.07 & 98.99 & $93.58\pm{0.21}$  & $91.44\pm{0.23}$ & $98.11\pm{0.09}$
 & $87.08\pm{0.15}$ & ${84.32\pm{0.52}}$ & $92.51\pm{0.51}$\\

\addlinespace
\hline
\addlinespace
$2^{nd}$ order Tree-RNN (\textbf{ours}) & 95.62 & 94.58 & 98.25 & $90.28 \pm{0.07}$  & $90.25\pm{0.35}$ & $98.08\pm{0.20}$
& $86.05\pm{0.11}$ & $84.05\pm{0.30}$ & $88.99\pm{0.30}$ \\
MI-Tree-LSTM  (\textbf{ours}) & 98.80 & 98.01 & 99.80 & $94.20\pm{0.02}$ & $91.50\pm{0.07}$ & $98.99\pm{0.06}$
 & $87.80\pm{0.7}$ & $84.08\pm{0.45}$ & $93.00\pm{0.40}$ \\
MTree-LSTM  (\textbf{ours}) & 98.25 & 97.00 & 98.81 & $94.09\pm{0.02}$ & $91.00\pm{0.05}$ & $99.00\pm{0.06}$
 & $87.25\pm{0.5}$ & $84.29\pm{0.25}$ & $93.02\pm{0.35}$ \\
MI-Tree-LSTM + stack  (\textbf{ours}) & 96.28 & 96.00 & 99.25 & $93.99\pm{0.04}$ & $91.55\pm{0.15}$ & $98.57\pm{0.09}$
 & $88.25\pm{0.81}$ & $84.41\pm{0.39}$ & $\mathbf{94.47\pm{0.61}}$ \\
M-Tree-LSTM + stack  (\textbf{ours}) & 97.35 & 97.15 & 99.20 & $94.29\pm{0.02}$ & $90.99\pm{0.04}$ & $98.28\pm{0.09}$
 & $\mathbf{89.04\pm{0.54}}$ & $\mathbf{84.59\pm{0.47}}$ & $94.39\pm{0.59}$ \\
\bottomrule
\end{tabular}}
\end{table*}

\begin{table*}[htb]
  \caption{\textbf{Examples of generated equations used in the paper's experiments \cite{arabshahi2019memory}}.}
\label{tab:examples}
\centering
\resizebox{1\textwidth}{!}{%
\begin{tabular}{llc}
\toprule
\bf Example & \bf Label & \bf Depth \\ 
\midrule
\addlinespace
$(\sqrt{1} \times 1 \times y) + x = (1 \times y) + x$ & Correct & 4 \\
\addlinespace
$\sec(x + \pi) = (-1 \times \sec(\sec(x)))$ & Incorrect & 4 \\
\addlinespace
$y \times \Big(1^1 \times (3 + (-1 \times 4^{0 \times 1})) + x^1\Big) = y \times 2^0 \times (2 + x)$ & Correct & 8 \\
\addlinespace
$\sqrt{1 + (-1 \times (\cos(y + x))^{\sqrt{\csc(2)}} )} \times (\cos(y + x))^{-1} = \tan(y^1 + x)$ & Incorrect & 8\\
\addlinespace
$2^{-1} + \Big(-\frac{1}{2} \times \big (-1 \times \sqrt{1 + (-1 \times \sin^2(\sqrt{4} \times (\pi + (x \times -1))))} \big) \Big) + \cos^{\sqrt{4}}(x) = 1$ & Correct & 13 \\
\addlinespace
 $\big(\cos(y^1 + x) + z\big)^w = \big(\cos(x) \times \cos(0 + y) + \big(-1 \times \sqrt{1 + -1 \times \cos^{2}(y + 2\pi)}\; \big) \times \sin(x) + z \big)^w$  & Correct & 13  \\
 \addlinespace
 $\sin\Big( \sqrt{4}\;^{-1} \pi + (-1 \times \sec \big( \csc^2(x)^{-1} + \sin^2(1 + (-1 \times 1) + x + 2^{-1} \pi ) \big) \times x) \Big) = \cos(0 + x)$ & Incorrect & 13 \\
\addlinespace
\bottomrule
\end{tabular}}
\end{table*}

\bibliography{aaai}